\documentclass{article}
\usepackage{graphicx}
\usepackage{hyperref}
\usepackage{dirtree}
\usepackage{subcaption}

\title{Badgers: generating data quality deficits with Python.}
\author{Julien Siebert, Daniel Seifert, Patricia Kelbert, Michael Kläs, Adam Trendowicz}
\date{\today}

\begin{document}

\maketitle

\begin{abstract}
Generating context specific data quality deficits is necessary to experimentally assess data quality of data-driven (artificial intelligence (AI) or machine learning (ML)) applications.
In this paper we present \texttt{badgers}, an extensible open-source Python library to generate data quality deficits (outliers, imbalanced data, drift, etc.) for different modalities (tabular data, time-series, text, etc.). The documentation is accessible at \url{https://fraunhofer-iese.github.io/badgers/} and the source code at \url{https://github.com/Fraunhofer-IESE/badgers}.
\end{abstract}

\section{Introduction}
\subsection{Context}
Applications and systems based on artificial intelligence (AI), machine learning (ML), data mining or statistics (hereafter referred to as data-driven software components) are pieces of software where the decision function is not programmed in a classical way, but is based on one or more models that can be designed either automatically (e.g. through learning or mining) or is based on domain expertise hypotheses (e.g. business rules or statistical tests).
Assessing the quality of such software components is not trivial, as it depends on several factors, such as the quality and quantity of the data, the type of model and how it is built, the application context, and domain expertise \cite{siebert2022construction}.
\subsection{Motivation}\label{sec:motivation}
Data quality deficits (e.g., outliers, imbalanced data, missing values, etc.) can have a variety of effects on the performance of a data-driven model. A theoretical understanding of the robustness of data-driven models against specific data quality deficits is available for only a small number of models. Many can only be empirically tested against specific data quality deficits. To make matters worse, data quality deficits are context and application dependent.
Assessing the robustness of a data-driven software components to changes in data quality requires a systematic approach. It also requires the ability to generate specific data quality deficits in order to run tests.
\paragraph{}
Currently, there are many Python libraries to detect and handle data quality deficits, such as pyod\footnote{\url{https://pyod.readthedocs.io/en/latest/}} \cite{zhao2019pyod} for detecting outliers, imbalanced-learn\footnote{\url{https://imbalanced-learn.org}} \cite{lemaitre2017imbalanced} for dealing with imbalanced data, autoimpute\footnote{\url{https://autoimpute.readthedocs.io/en/latest/}} for imputing  missing values, or great-expectations\footnote{\url{http://docs.greatexpectations.io}} for validation. In addition, the field of deep-learning has provided us with libraries for augmenting training data (see for instance albumentation\footnote{\url{https://albumentations.ai/docs/}} \cite{buslaev2020albumentation}). However, there are very few, if any, libraries for generating context-specific data quality deficits.
\subsection{Contribution}
This paper presents \texttt{badgers}, a Python package dedicated to generate data quality deficits. The aim is to propose a set of standardized and extensible objects (called generators) that can take data as input, infer context information from it, and generate data quality deficits. This package relies on a few design decisions. First, it follows a simple API. Each generator provides a \texttt{generate(X, y)} function (where X is the input features and y is either a vector of class labels, regression targets, or an empty one).  Secondly, \texttt{badgers} aims to support as many data types as possible (e.g., tabular data, images, text, graphs, etc.). This means relying on mainstream and long-established libraries (such as numpy\footnote{\url{https://numpy.org/}}, pandas\footnote{\url{https://pandas.pydata.org/}, or scikit-learn\footnote{\url{https://scikit-learn.org/stable/index.html}}} for tabular data) whenever possible, or following reasonable design decisions. Finally, \texttt{badgers} should be structured and implemented so that it can be easily extended. 
\subsection{Structure of the paper}
The paper is organized as follows. Section \ref{sec:related-work} presents a short overview of related work. Section \ref{sec:proposed-solution} presents \texttt{badgers} structure and implementation. Section \ref{sec:examples} shows a couple of application examples. Section \ref{sec:conclusion} discusses limitations, future work, concludes the paper and provides links to the project.

\section{Related Work\label{sec:related-work}}
Assessing the quality of ML applications is a broad area of research. In their paper \cite{zhang2020machine}, Zhang and co-authors provide a relatively comprehensive overview of testing activities that apply to machine learning. According to their categorization, we can argue that generating data quality defects falls into the spectrum of test input generation. That is, the generation of specific data with the purpose of evaluating specific aspects of the system under test. The techniques listed range from rule-based to generative AI techniques. Most of the methods presented here are either part of specific test frameworks or have been described in scientific papers. To the best of our knowledge, they are not part of a library dedicated to the generation of quality defects.

Data augmentation techniques are typically used in machine learning to enrich the training data set and help train models to achieve a better goodness of fit, generalize better, and become robust to some data quality issues (e.g., noise). They usually consist of specific transformations (like rotations or scaling for images) that, in principle, should not change the semantic of the data. Recent surveys, like  \cite{shorten2019survey,kaur2021data} for images and \cite{bayer2022survey,shorten2021text} for text, provide an overview of the different techniques used in data augmentation. In section \ref{sec:motivation}, we mentioned existing libraries for data augmentation. Although their main goal is not to specifically generate data quality deficits, data augmentation methods provide interesting algorithms that can be reused for our purpose.

When it comes to generating data quality deficits from existing data, very few papers provide overviews of existing methods and implementations. For instance, \cite{steinbuss2021outliers} discusses how to generate outliers from existing data. While the authors seem to have implemented a number of these methods to test them empirically, no implementation is actually available.
\cite{santos2019missing} discusses how to generate missing values. Note that the methods discussed in \cite{santos2019missing} have been implemented in R\footnote{ \url{https://cran.r-project.org/web/packages/missMethods/}} but not in Python.

In summary, there exists a variety of methods for generating data quality defects. But very few are available in a dedicated Python library.

\section{Proposed solution: Badgers\label{sec:proposed-solution}}
\subsection{Overview}
Badgers is a Python library for generating data quality deficits from existing data. As a basic principle, badgers provides a set of objects called generators that follow a simple API: each generator provides a \texttt{generator(X,y)} function that takes as argument \texttt{X} (the input features) and \texttt{y} (the class labels, the regression target, or None) and returns the corresponding transformed \texttt{Xt} and \texttt{yt}. As an example, figure \ref{fig:example_generate} shows the generate function implemented in the \texttt{Gaussian\-Noise\-Generator} that adds Gaussian White noise to some existing data.

\begin{figure}[!htbp]
    \begin{verbatim}
def generate(self, X, y, **params):
    """
    Add Gaussian white noise to the X.
    X is first standardized (each column has a mean = 0 and variance = 1).
    Noise is generated from a normal distribution with 
    standard deviation = `self.noise_std`.
    Noise is added to X.

    :param X: the input
    :param y: the target
    :param params: optional parameters
    :return: Xt, yt
    """
    # standardize X
    scaler = StandardScaler()
    # fit, transform
    scaler.fit(X)
    Xt = scaler.transform(X)
    # add noise
    Xt = Xt + self.random_generator.normal(
        loc=0, 
        scale=self.noise_std, 
        size=Xt.shape
    )
    # inverse pca
    return scaler.inverse_transform(Xt), y
    \end{verbatim}
    \caption{Generate function implemented in the \texttt{Gaussian\-Noise\-Generator}. This function first standardize the input features \texttt{X}. It then adds Gaussian White noise to it. Finally it returns the noisy data in the original feature space (by inverting the standardization).}
    \label{fig:example_generate}
\end{figure}

\begin{figure}[!htbp]
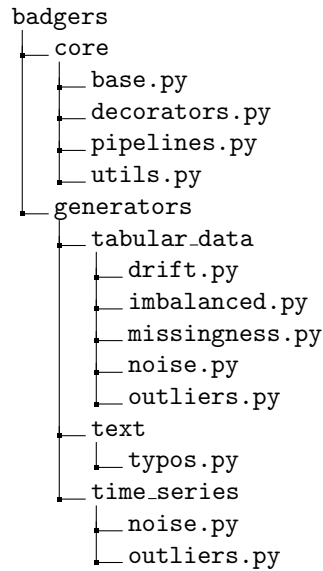

    \dirtree{%
        .1 badgers.
        .2 core.
        .3 base.py.
        .3 decorators.py.
        .3 pipelines.py.
        .3 utils.py.
        .2 generators.
        .3 tabular\_data.
        .4 drift.py.
        .4 imbalanced.py.
        .4 missingness.py.
        .4 noise.py.
        .4 outliers.py.
        .3 text.
        .4 typos.py.
        .3 time\_series.
        .4 noise.py.
        .4 outliers.py.
    }
    \caption{Current structure of the package.}
    \label{fig:badgers-tree}
\end{figure}

The code is divided into two main modules: \texttt{core} and \texttt{generators}. The \texttt{core} module handles all the utilities and things that are generic to all generators such as base classes (in \texttt{base.py}), decorators (in \texttt{decorators.py}), and utilities (in \texttt{utils.py}). The generators themselves are stored under the \texttt{generators} module, which in turn is divided into submodules, each representing a data type (e.g. \texttt{tabular\_data}, \texttt{time\_series}, \texttt{images}, \texttt{graphs}, etc.). Each submodule hosts the generators implementations dedicated to one specific data quality deficit (such as outliers, drift, missingness, etc.) for a specific data type. Figure \ref{fig:badgers-tree} shows the detail of the current structure.

\subsection{Available features}
Badgers is currently under development and the list of features will most probably evolve in the near future. For the moment, the focus has been more on tabular data. As shown in the figure \ref{fig:badgers-tree}, the module \texttt{bad\-gers.gen\-er\-at\-ors.tabular\_data} contains five submodules: \texttt{drift.py}, \texttt{imbalanced.py}, \texttt{mis\-sing\-ness.py}, \texttt{noise.py}, and \texttt{outliers.py}. As their names suggest, each submodule implements generators dedicated to specific data quality deficits. For time series data (\texttt{bad\-gers.gen\-er\-at\-ors.time\_series}), the following submodules are available: \texttt{noise.py} and \texttt{outliers.py}. For text data (\texttt{bad\-gers.gen\-er\-at\-ors.time\_series}) only one submodule (\texttt{typos.py}) is at the moment available.
\subsubsection{Tabular Data}
\begin{figure}[!htbp]
    \centering
    \includegraphics[width=\textwidth]{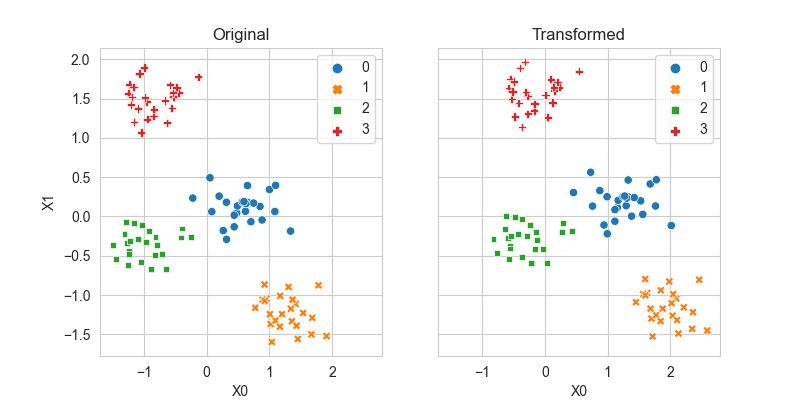}
    \caption{\texttt{RandomShiftGenerator}: the whole data set is translated at once.}
    \label{fig:drift-random}
\end{figure}

\begin{figure}[!htbp]
    \centering
    \includegraphics[width=\textwidth]{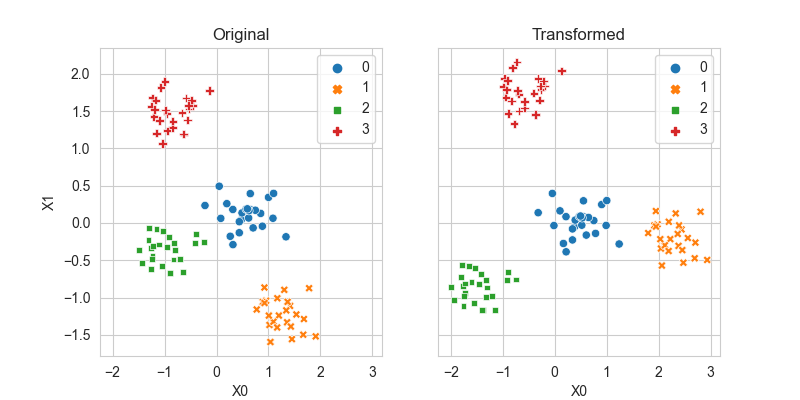}
    \caption{\texttt{RandomShiftClassesGenerator}: each class is translated independently.}
    \label{fig:drift-random-classes}
\end{figure}

\paragraph{bad\-gers.gen\-er\-at\-ors.tabular\_data.drift}
Drift happens when some statistical properties of the data changes over time \cite{lu2018learning}. Two generators are currently available in this module: \texttt{Random\-Shift\-Transformer} and \texttt{Random\-Shift\-Classes\-Transformer}. Figures \ref{fig:drift-random} and \ref{fig:drift-random-classes} illustrate how these two generators works. Simply put, the \texttt{Random\-Shift\-Transformer} randomly shifts values of each column independently of one another. This amounts to translating the data (see Figure \ref{fig:drift-random}). The input features are first standardized (mean = 0, var = 1) and a random number is added to each column. The \texttt{Random\-Shift\-Classes\-Transformer} applies a similar transformation but for instances belonging to the same class. Here all the instances of a given class are translated, and the translation for different classes is not the same (see Figure \ref{fig:drift-random-classes}).

\begin{figure}[!htbp]
    \centering
    \includegraphics[width=\textwidth]{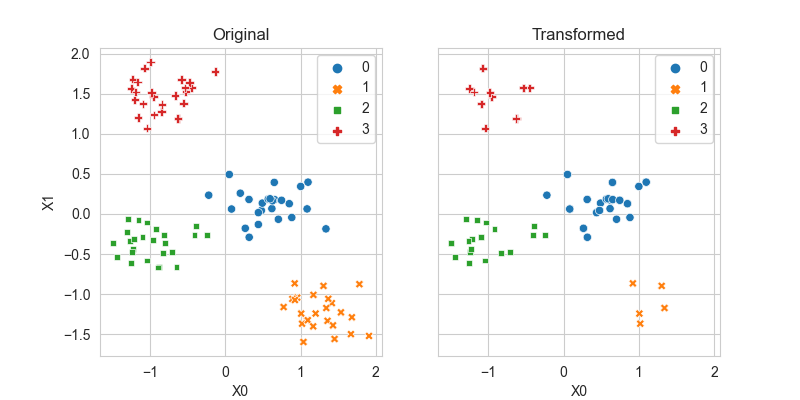}
    \caption{\texttt{Random\-Sampling\-Classes\-Generator}: this is the classical example of imbalanced data where some classes are under- and some classes are over-represented.}
    \label{fig:random-sampling-classes}
\end{figure}

\begin{figure}[!htbp]
    \centering
    \includegraphics[width=\textwidth]{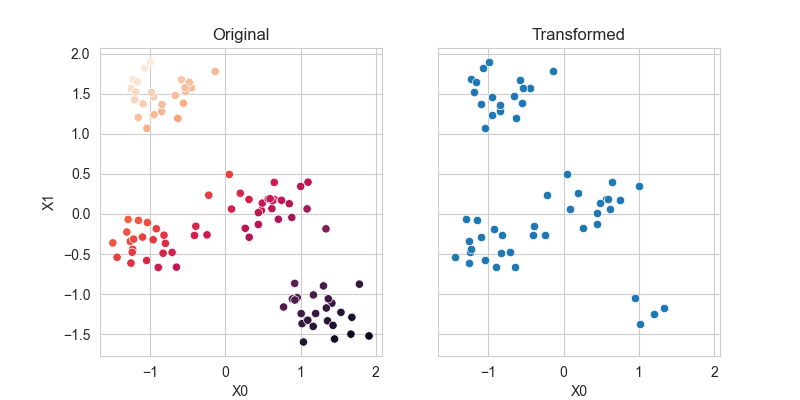}
    \caption{\texttt{Random\-Sampling\-Targets\-Generator}: here the probability of a data point being removed depends on the regression target. This is represented by the color gradient on the left (the darker the color, the higher the probability of being removed).}
    \label{fig:random-sampling-target}
\end{figure}

\begin{figure}[!htbp]
    \centering
    \includegraphics[width=\textwidth]{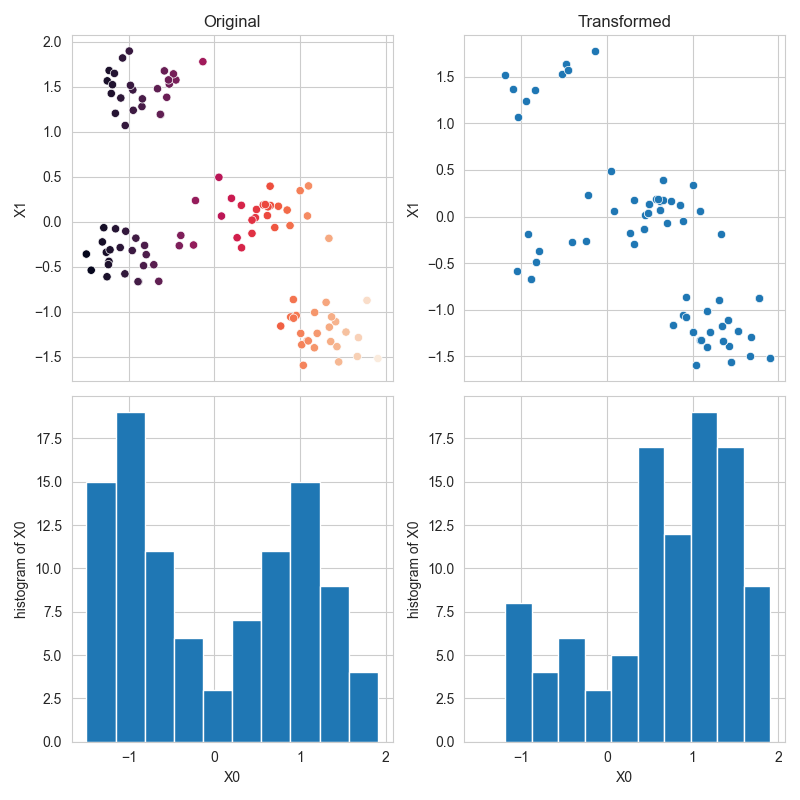}
    \caption{\texttt{Random\-Sampling\-Features\-Generator}: in this example, the probability of being removed is set to be proportional to the first feature ($X0$). As in the previous figure, the color gradient in the upper left figure represents the probability of a data point being removed (the darker the color, the higher the probability of being removed). The two histogram below represents the distribution of the $X0$ values (i.e., the marginal probability distribution of $X0$).}
    \label{fig:random-sampling-feature}
\end{figure}

\paragraph{bad\-gers.gen\-er\-at\-ors.tabular\_data.imbalanced}
Whereas imbalanced data is usually understood in the context of classification \cite{he2009learning}, when some classes are over- or under-represented, we use a broader definition. For us, a data set is said to be imbalanced when some statistical properties of the data are over- or under-represented in comparison to a ground truth. Currently, three generators have been implemented: \texttt{Random\-Sampling\-Classes\-Generator}, \texttt{Random\-Sampling\-Targets\-Generator}, and \texttt{Random\-Sampling\-Features\-Generator}. Simply put, all of these generators sample the original data set with replacement. The \texttt{Random\-Sampling\-Classes\-Generator} samples data points belonging to each class to obtain a specified class distribution (e.g., $10\%$ of class 1, $20\%$ of class 2, and $70\%$ of class 3, see Figure \ref{fig:random-sampling-classes}). The \texttt{Random\-Sampling\-Targets\-Generator} samples data points according to the regression target \texttt{y} and expects a function that maps the values of \texttt{y} to a sampling probability (see Figure \ref{fig:random-sampling-target}). Finally, the \texttt{Random\-Sampling\-Features\-Generator} performs a similar transformation but the sampling probability now depends upon the input features values \texttt{X} (see Figure \ref{fig:random-sampling-feature}).

\begin{figure}[!htbp]
    \centering
    \includegraphics[width=\textwidth]{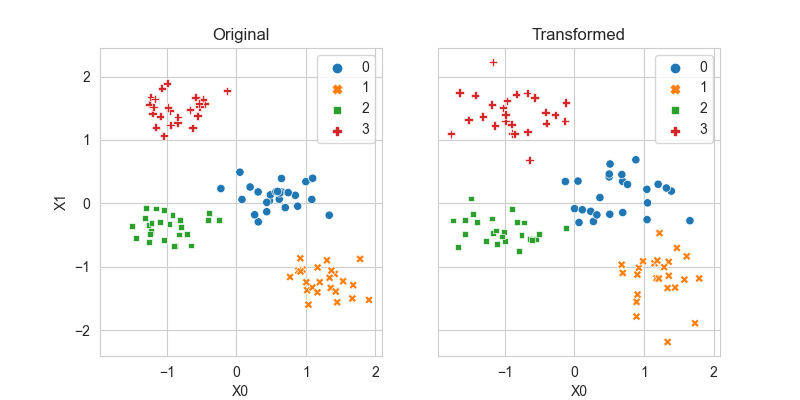}
    \caption{\texttt{Gaussian\-Noise\-Generator}: this adds Gaussian white noise to all data points.}
    \label{fig:gauss_noise}
\end{figure}

\newpage
\paragraph{bad\-gers.gen\-er\-at\-ors.tabular\_data.noise}
Currently only one generator has been implemented: \texttt{Gaussian\-Noise\-Generator}. It adds a Gaussian White noise to the input features \texttt{X} (see Figure \ref{fig:gauss_noise}).

\paragraph{bad\-gers.gen\-er\-at\-ors.tabular\_data.outliers}
Two types of generators are currently available. Generators that directly generate outliers from the input features \texttt{X} and generators that first reduce the dimensionality of the input features \texttt{X} and then apply an outlier generator from the previous category. 

\begin{figure}[!htbp]
    \centering
    \includegraphics[width=0.75\textwidth]{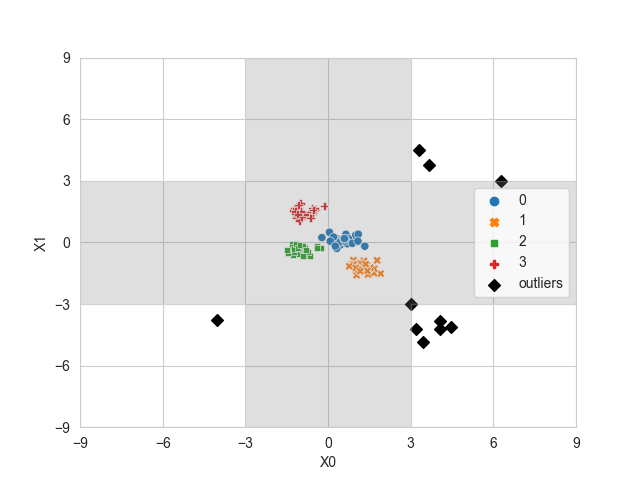}
    \caption{\texttt{ZScoreSamplingGenerator}: each feature of the outliers generated lies outside the range $]\mu_i-3\sigma_i,\mu_i+3\sigma_i[$. This corresponds to the white regions. Note that in this example the data is standardized ($\mu_i=0$ and $\sigma_i=1$ $\forall i\in[0,1]$).}
    \label{fig:outliers-zscore}
\end{figure}

\begin{figure}[!htbp]
    \centering
    \includegraphics[width=0.75\textwidth]{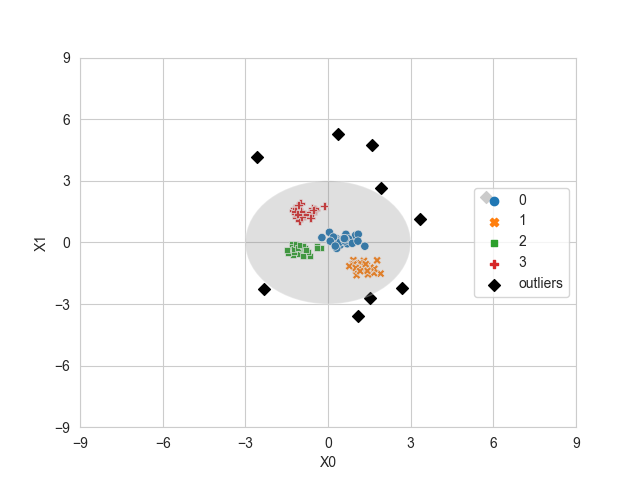}
    \caption{\texttt{HypersphereSamplingGenerator}: outliers are generated outside the grey disc centered on $\mu$ with radius $\geq 3\sigma$. In this example, the data is standardized ($\mu_i=0$ and $\sigma_i=1$ $\forall i\in[0,1]$).}
    \label{fig:outliers-hypersphere}
\end{figure}

\begin{figure}[!htbp]
    \centering
    \includegraphics[width=0.75\textwidth]{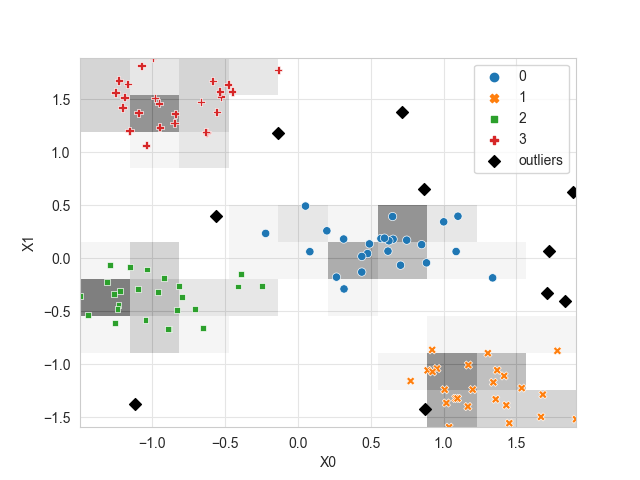}
    \caption{\texttt{HistogramSamplingGenerator}: outliers are generated in regions of low density. These regions are approximated by a histogram. The shade of grey represents the density: the darker the region, the denser it is. Therefore, outliers should be sampled in a region with a lighter shade.}
    \label{fig:outliers-histogram}
\end{figure}

\begin{figure}[!htbp]
    \centering
    \includegraphics[width=0.75\textwidth]{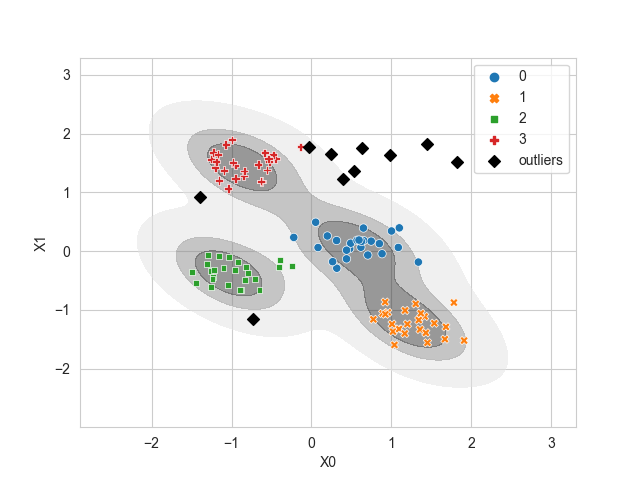}
    \caption{\texttt{LowDensitySamplingGenerator}: similar to the previous figure, outliers are generated in regions of low density. These regions are approximated by a \texttt{KernelDensityEstimator} from scikit-learn. The shade of grey represents the density: the darker the region, the denser it is. Therefore, outliers should be sampled in a region with a lighter shade.}
    \label{fig:outliers-lowdensity}
\end{figure}

\texttt{ZScore\-Sampling\-Generator}, \texttt{Hypersphere\-Sampling\-Generator}, \texttt{Histogram\-Sampling\-Generator}, and \texttt{Low\-Density\-Sampling\-Generator} are generators from the first category. Figures \ref{fig:outliers-zscore}, \ref{fig:outliers-hypersphere}, \ref{fig:outliers-histogram}, and \ref{fig:outliers-lowdensity} illustrate how these four generators create outliers.

The \texttt{ZScore\-Sampling\-Generator} generates outliers by creating data points where each feature $i$ gets a value outside the range $]\mu_i-3\sigma_i,\mu_i+3\sigma_i[$, where $\mu_i$ and $\sigma_i$ are the mean and the standard deviation of feature $i$ (see Figure \ref{fig:outliers-zscore}). The \texttt{Hypersphere\-Sampling\-Generator} generates outliers by creating data points on an hypersphere of center $\mu$ and of radius larger than $3\sigma$ (see Figure \ref{fig:outliers-hypersphere}). The \texttt{Histogram\-Sampling\-Generator} and \texttt{Low\-Density\-Sampling\-Generator} both generate outliers by creating data points that belong to regions of low density. The difference between the two generators lies in their low density estimation methods. \texttt{Histogram\-Sampling\-Generator} approximates regions of low density by computing an histogram of the data (see figure \ref{fig:outliers-histogram}). \texttt{Low\-Density\-Sampling\-Generator} uses a kernel density estimator (see Figure \ref{fig:outliers-lowdensity}).

\texttt{Decomposition\-And\-Outlier\-Generator} belongs to the second category. It first standardizes the data and applies a dimensionality reduction technique (so far badgers support scikit-learn transformers that provide an \texttt{inverse\_transform} function like \texttt{sklearn.decomposition.PCA}). The outliers are generated using one of the generators mentioned above. Finally the standardization and the dimensionality reduction are inverted.

\subsubsection{Time series data}
\begin{figure}[!htbp]
    \centering
    \includegraphics[width=\textwidth]{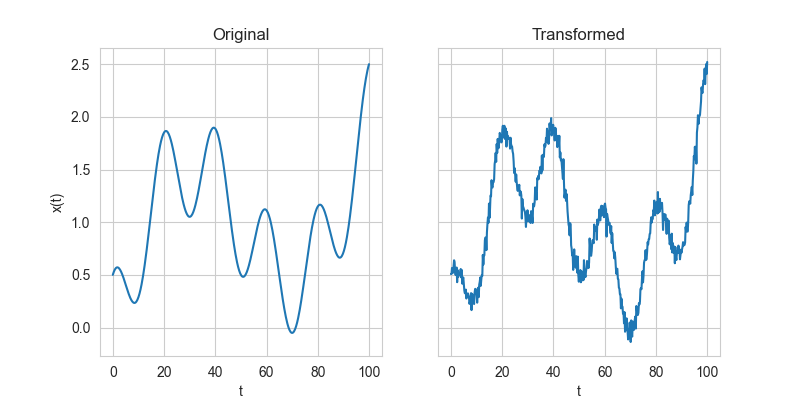}
    \caption{\texttt{Gaussian\-Noise\-Generator}: this generator adds Gaussian white noise to the original signal.}
    \label{fig:time-series-noise}
\end{figure}

Time series data is currently supported in \texttt{badgers} in the form of numpy arrays and pandas dataframes. 
\paragraph{bad\-gers.gen\-er\-at\-ors.time\_series.noise}
Currently only one generator has been implemented: \texttt{Gaussian\-Noise\-Generator}. It adds a Gaussian White noise to the input features \texttt{X}. The implementation is the same as in \texttt{bad\-gers.gen\-er\-at\-ors.tabular\_data.noise}. Figure \ref{fig:time-series-noise} illustrates this generator.

\begin{figure}[!htbp]
    \centering
    \includegraphics[width=\textwidth]{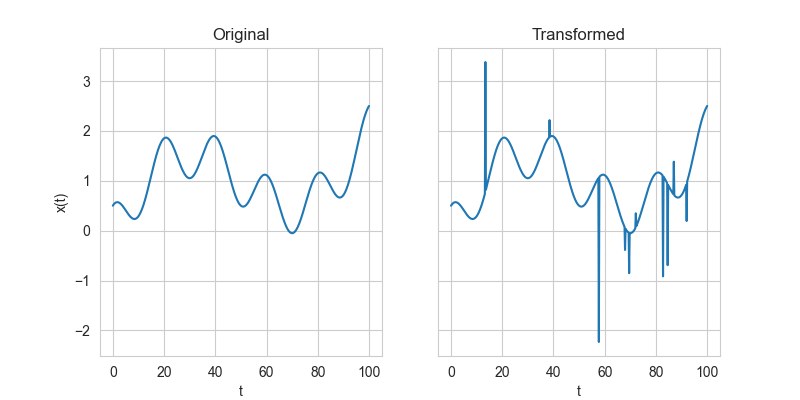}
    \caption{\texttt{Local\-ZScore\-Generator}: local extreme values are added to the original signal.}
    \label{fig:local-zscore}
\end{figure}

\paragraph{bad\-gers.gen\-er\-at\-ors.time\_series.outliers}
Here some existing instances are replaced with outliers. Currently only one generator is implemented: \texttt{Local\-ZScore\-Generator}. The \texttt{LocalZ\-Score\-Generator} creates locally extreme values, by changing the values of some randomly selected data points $x(t_i) \in X$ (see Figure \ref{fig:local-zscore}). The values are sampled out of the $]\mu_{j,\Delta}-3\sigma_{j,\Delta},\mu_{j,\Delta}+3\sigma_{j,\Delta}[$ range, where $\mu_{j,\Delta}$ and $\sigma_{j,\Delta}$ are the mean and the standard deviation of the $j^{th}$ feature computed in the local time interval $\Delta = [t_i - n, t_i + n]$.

\subsubsection{Text}
Text data is currently supported in \texttt{badgers} in the form of lists of strings. 
\paragraph{bad\-gers.gen\-er\-at\-ors.text.typos}
For now, only one generator is implemented: \texttt{Swap\-Letters\-Generator}. The \texttt{Swap\-Letters\-Generator} randomly swaps adjacent letters in words larger than three letters except for the first and the last letters. As an illustration, the sentence "the quick brown fox jumps over the lazy dog" becomes "the qucik brwon fox jupms oevr the lzay dog" after applying this generator.

\section{Examples\label{sec:examples}}
We implemented several examples in the form of notebooks (accessible at \url{https://fraunhofer-iese.github.io/badgers/} under the tutorials section). The next two figures provide some examples to illustrate the use of a single generator (Figure \ref{fig:code_example_imbalanced}), as well as the pipelining of several ones (Figure \ref{fig:code_example_pipeline}).

\begin{figure}[!htpb]
    \begin{verbatim}
from sklearn.datasets import make_blobs
from badgers.generators.tabular_data.imbalance import \
    RandomSamplingClassesGenerator
from badgers.core.utils import normalize_proba
# load data
X, y = make_blobs(centers=4, random_state=0)

trf = RandomSamplingClassesGenerator(
    proportion_classes={0:0.5, 1:0.05, 2:0.25, 3:0.2}
)
Xt, yt = trf.generate(X.copy(),y)
    \end{verbatim}
    \caption{A code example showing how to generate an imbalanced data set by randomly sampling data points of different classes. The existing data is generated using the \texttt{make\_blob} function from \texttt{scikit-learn}. It contains 4 classes. The \texttt{proportion\_classes} variable is used for determining how many data points from each class will be present in the resulting imbalanced data set.}
    \label{fig:code_example_imbalanced}
\end{figure}

\begin{figure}[!htbp]
    \begin{verbatim}
from sklearn.datasets import make_blobs
from badgers.generators.tabular_data.noise import \
    GaussianNoiseGenerator
from badgers.generators.tabular_data.imbalance import \
    RandomSamplingClassesGenerator
from badgers.core.pipeline import Pipeline

X, y = make_blobs(centers=4, random_state=0)

generators = {
    'imbalance': RandomSamplingClassesGenerator(
        proportion_classes={0:0.6, 1:0.25, 2:0.1, 3:0.05}
    ),
    'noise': GaussianNoiseGenerator(noise_std=0.5)
}

pipeline = Pipeline(generators=generators)

Xt, yt = pipeline.generate(X.copy(),y)
    \end{verbatim}
    \caption{A code example illustrating the pipelining of several generators. The data is first sampled for imbalancedness, then noise is added.}
    \label{fig:code_example_pipeline}
\end{figure}

\newpage
\section{Conclusion\label{sec:conclusion}}
This paper gave an overview of \texttt{badgers}, a Python package dedicated to generating data quality deficits. 

\texttt{Badgers} is in a relatively early development stage. Until now, our focus has been to develop the library structure, the API, as well as some relatively simple generators. The goal was first and foremost to show the potential of such a library.

This library has been used in the context of internal projects. The purpose was first to conduct robustness tests and to augment data. By open-sourcing this library, we hope to provide not only a tool to ease robustness tests of data-driven applications but also to foster discussions on the topic of generating data quality deficits.

Future work will focus both on developing new generators and to test the applicability of this library in the context of data science projects. Discussions and design decisions will be needed to prioritize the work and to decide how to improve the support of other types of data (for instance images, graphs, geolocated data).

Finally, \texttt{badgers} can be installed with the Python package installer \texttt{pip}\footnote{\url{https://pip.pypa.io/en/stable/}}: \texttt{pip install badgers}.
The full documentation is accessible at \url{https://fraunhofer-iese.github.io/badgers/}.
The source code for \texttt{badgers} is available under the BSD-3 license at \url{https://github.com/Fraunhofer-IESE/badgers}.

\bibliographystyle{alpha}
\bibliography{main}
\end{document}